\documentclass[10pt,twocolumn,letterpaper]{article}

\usepackage{cvpr}              %

\definecolor{cvprblue}{rgb}{0.21,0.49,0.74}
\usepackage[pagebackref,breaklinks,colorlinks,allcolors=cvprblue]{hyperref}

\usepackage{makecell}
\usepackage{graphicx}
\usepackage{amsmath, bm}
\usepackage{tabularx,booktabs}
\usepackage{adjustbox}
\usepackage{multirow}
\usepackage{graphicx,xcolor,colortbl} 
\definecolor{LightCyan}{rgb}{0.88,1,1}
\usepackage{array}
\newcolumntype{C}[1]{>{\centering\arraybackslash}p{#1}}
\newcolumntype{L}[1]{>{\arraybackslash}p{#1}}

\usepackage{subcaption}
\usepackage{indentfirst}
\usepackage{pifont}
\usepackage{lipsum}
\usepackage[symbol]{footmisc}
\usepackage[dvipsnames]{xcolor}

\usepackage{pifont}
\newcommand{\cmark}{\ding{51}}

\DeclareMathAlphabet{\mathsfit}{\encodingdefault}{\sfdefault}{m}{sl}
\SetMathAlphabet{\mathsfit}{bold}{\encodingdefault}{\sfdefault}{bx}{n}

\newcommand{\R}{\mathbb{R}}

\newcommand{\inputset}{\mathcal{I}}  %
\newcommand{\inputimage}{\mathbf{I}}  %
\newcommand{\inputheadset}{\mathbf{H}}  %
\newcommand{\jointnum}{\mathcal{J}}  %
\newcommand{\rotatenum}{\mathcal{K}}
\newcommand{\featuremap}{\mathbf{F}}  %

\newcommand{\outputpose}{p}  %
\newcommand{\uncertainty}{s}  %
\newcommand{\studenttd}{d}  %
\newcommand{\studenttv}{\nu}  %
\newcommand{\cholesky}{\mathbf{L}}  %
\newcommand{\covariance}{\bm{\Sigma}}  %
\newcommand{\residual}{\bm{\delta}}  %

\def\paperID{10986} %
\def\confName{CVPR}
\def\confYear{2026}

\newcommand{\ourmethodtitle}{EgoPoseFormer v2}
\newcommand{\ourmethod}{EPFv2\xspace}

\title{EgoPoseFormer v2: Accurate Egocentric Human Motion Estimation for AR/VR}

\author{
Zhenyu Li$^{12}$\footnotemark[2], Sai Kumar Dwivedi$^{13}$\footnotemark[2], Filip Maric$^1$, Carlos Chacón$^1$, \\
Nadine Bertsch$^1$, Filippo Arcadu$^1$, Tomas Hodan$^1$, Michael Ramamonjisoa$^1$,\\
Peter Wonka$^2$, Amy Zhao$^1$, Robin Kips$^1$, Cem Keskin$^1$, Anastasia Tkach$^1$\footnotemark[1], Chenhongyi Yang$^1$\footnotemark[1]\\
\\
$^1$Meta, $^2$KAUST, $^3$Max Planck Institute for Intelligent Systems  \\
}

\begin{document}

\twocolumn[{%
\renewcommand\twocolumn[1][]{#1}%
\maketitle
\vspace{-35pt}
\begin{center}
    \centering
    \captionsetup{type=figure}
    \includegraphics[width=0.99\textwidth]{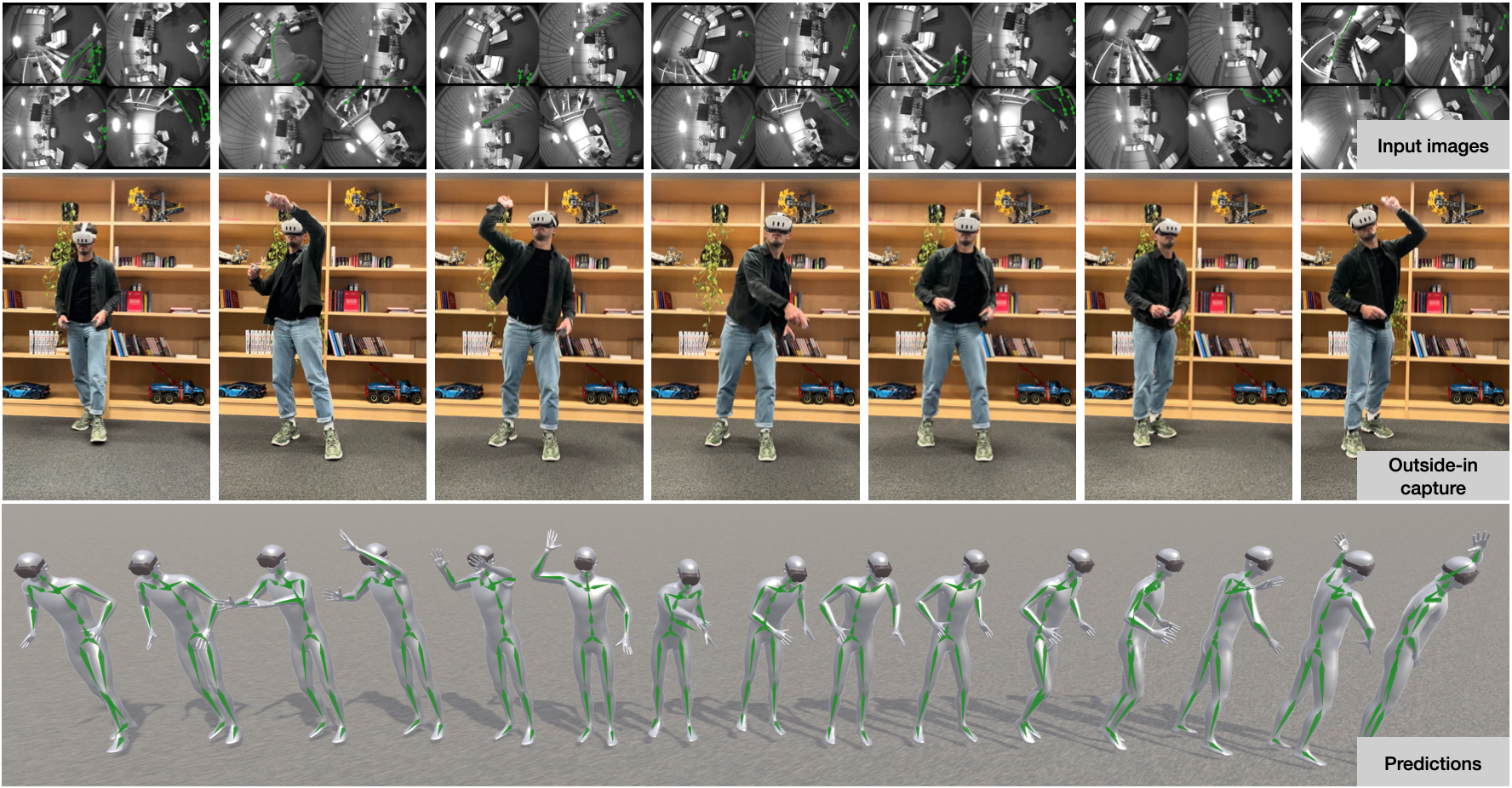}
    \vspace{-6pt}
    \captionof{figure}{\textbf{Results on an in-the-wild sequence.} Top to bottom: (1) input egocentric images with 2D projections of estimated body joints, (2) outside-in capture, and (3) renderings of estimated body motion. \ourmethod demonstrates accurate and temporally-consistent estimates.
    }
    \label{fig:teaser}
\end{center}%
}]

\footnotetext[1]{Equal supervision.}
\footnotetext[2]{Work done during internship at Meta.}

\begin{center}
\section*{Abstract}
\end{center}
{\it
Egocentric human motion estimation is essential for AR/VR experiences, yet remains challenging due to limited body coverage from the egocentric viewpoint, frequent occlusions, and scarce labeled data. We present \ourmethodtitle, a method that addresses these challenges
through two key contributions: (1)~a transformer-based model for temporally consistent and spatially grounded body pose estimation, and (2)~an auto-labeling system that enables the use of large unlabeled datasets for training.
Our model is fully differentiable, introduces identity-conditioned queries, multi-view spatial refinement, causal temporal attention, and supports both keypoints and parametric body representations under a constant compute budget.
The auto-labeling system scales learning to tens of millions of unlabeled frames via uncertainty-aware semi-supervised training. The system follows a~teacher–student schema to generate pseudo-labels and guide training with uncertainty distillation, enabling the model to generalize to different environments. On the EgoBody3M benchmark, with a 0.8 ms latency on GPU, our model outperforms two state-of-the-art methods by 12.2\% and 19.4\% in accuracy, and reduces temporal jitter by 22.2\% and 51.7\%. Furthermore, our auto-labeling system further improves the wrist MPJPE by 13.1\%.  
}
    
\vspace{10pt}
\section{Introduction}
\label{sec:intro}

Egocentric 3D motion estimation is an essential task for AR/VR, enabling input and interaction with virtual content~\cite{applevisionpro,quest3,plizzari2024outlook}.
However, recovering full 3D body motion solely from head-mounted cameras remains challenging.
The egocentric viewpoint offers only limited coverage of the body, suffers from frequent self-occlusions, and provides limited scene context~\cite{azam2024survey,li2025challenges}.
Moreover, egocentric data are often device-specific and sparsely annotated, making large-scale supervised training infeasible. 
As a result, existing pose estimation methods often produce inaccurate and jittery motion estimates and generalize poorly to new devices and in-the-wild scenarios.

Despite recent improvements in egocentric motion estimation, most existing methods still struggle to produce smooth and plausible poses when the body is occluded or out of sight.
Early approaches like EgoGlass and UnrealEgo~\cite{akada2022unrealego,kang2023ego3dpose,zhao2021egoglass,tome2019xregopose} rely on single-frame heatmap regression, often leading to poor accuracy~\cite{yang2024epfv1} and temporal inconsistencies.
LSTM-based EgoBody3M~\cite{zhao2024egobody3m} improve temporal smoothness but lack explicit modeling of 3D geometric relationships between joints and image features. 
EgoPoseFormer~\cite{yang2024epfv1}, a more recent transformer-based model improves 3D accuracy by projecting estimated 3D joints onto 2D images and using these projections to guide where features are extracted from each camera view via deformable attention.
However, this design uses a separate representation for each joint (scaling linearly with each joint), and the two-stage architecture prevents gradients from flowing back to the initial coarse pose estimate, limiting end-to-end learning.
Another approach~\cite{akada2024unrealego2} incorporates temporal information by conditioning the initial pose estimate on temporal features and applying self-attention modules, but it requires additional depth and segmentation modules, limiting the real-time applicability.

We introduce \textit{\ourmethodtitle~(\ourmethod)}, motivated by recent advances in foundation models~\cite{radford2021clip,rombach2022sd,kirillov2023sam,wang2025vggt,simeoni2025dinov3} driven by two key factors: effective model design and scalable training data.
\ourmethod~tackles the challenging egocentric motion estimation by two distinct innovations: 1)~an end-to-end transformer architecture designed for joint spatio-temporal reasoning, and 2)~a scalable semi-supervised training approach that leverages large-scale unlabeled egocentric videos through automatic labeling. These two components are complementary and together enable accurate and temporally-consistent motion estimation.

The \ourmethod model is a fully differentiable, end-to-end transformer architecture.
Unlike EgoPoseFormer~\cite{yang2024epfv1} that uses separate learnable representation (query) for each joint, \ourmethod uses a single holistic query, which is conditioned on pose-dependent metadata like user identity and headset pose, to represent the entire body state.
This design makes computation independent to the body representation being predicted, improving efficiency and flexibility.
Our model consists of two architecturally identical transformer decoder blocks with full gradient flow between stages.
The first decoder predicts an initial 3D pose from temporal multi-view features.
The second decoder refines this estimation by projecting the coarse 3D keypoints onto image planes and using these 2D positions as spatial conditioning for cross-attention, guiding where to extract refinement features.
This achieves accurate geometric reasoning without computationally expensive deformable attention blocks~\cite{yang2024epfv1}.
Coupled with causal temporal attention, \ourmethod\ produces accurate, temporally-consistent motion even for invisible body parts, e.g., those under occlusion.

In addition to our model design, we propose a scalable semi-supervised learning pipeline that effectively leverages large collections of unlabeled egocentric videos.
While prior work has mitigated data scarcity using synthetic renderings or weak supervision~\cite{wang2022egopw,akada2022unrealego,akada2024unrealego2}, such approaches remain limited in realism or scale. 
Instead, \ourmethod leverages unlabeled in-the-wild data through semi-supervised learning, preserving real-world fidelity while drastically reducing labeling cost. 
Specifically, we train teacher model on a small labeled subset and then use it to generate pseudo labels for a large pool of unlabeled data~\cite{chapelle2009ssl,kirillov2023sam,simeoni2025dinov3,yang2024dav2}. 
The student is then trained jointly on labeled and pseudo-labeled samples using an uncertainty-guided distillation loss, which enables the student to recognize and down-weight unreliable pseudo-labels. This design makes EPFv2's performance scalable to the size of unlabeled data.

The resulted method achieves accurate, temporally-consistent, and generalizable human motion estimation. On the EgoBody3M benchmark~\cite{zhao2024egobody3m}, \ourmethod\ demonstrates strong quantitative and qualitative gains, highlighting the benefits of combining expressive model design with scalable data learning. 
Specifically, \ourmethod achieves a mean per-joint position error (MPJPE) of 4.02cm, representing a 22.4\% and 15.4\% improvement over EgoBody3M~\cite{zhao2024egobody3m} and EgoPoseFormer~\cite{yang2024epfv1}, respectively. \ourmethod\ also significantly improves temporal consistency, reducing MPJVE by 22.2\% and 51.7\% compared to the same baselines. Finally, with a full-model 0.8 ms GPU latency, our model is perfectly suitable for real-time VR devices. In summary, we make the following contruibutions:
\begin{enumerate}
    \item A transformer-based architecture that is end-to-end differentiable and performs efficient spatio–temporal reasoning from multi-view cameras.
    \item A scalable semi-supervised pipeline that leverages large unlabeled egocentric videos through uncertainty-guided teacher-student training.
    \item Extensive quantitative and qualitative evaluation demonstrates significant gains in accuracy, temporal consistency, and generalization on EgoBody3M benchmark.
\end{enumerate}

\section{Related work}
\label{sec:related}

\subsection{Egocentric motion estimation}

Early approaches for estimating egocentric pose relied primarily on heatmap-based methods~\cite{akada2022unrealego,kang2023ego3dpose,zhao2021egoglass,tome2019xregopose,xu2019mo2cap2,tome2020selfpose,wang2023scene,zhao2021egoglass}, where a dual-branch architecture predicts 2D joint heatmaps and regresses to 3D pose. While effective in constrained settings, these methods lack temporal consistency and suffer from jitter and instability during motion. To address this issue, EgoSTAN~\cite{park2022egotan} incorporated temporal context into the visual encoder, whereas EgoBody3M~\cite{zhao2024egobody3m} proposed an LSTM-based model to leverage sequence-level multi-view features for smoother predictions. More recently, transformer-based models~\cite{yang2024epfv1,akada2025egorear,akada2024unrealego2,camiletto2025frame} have been adopted for egocentric motion estimation. These approaches reformulate the task as a direct set prediction problem~\cite{carion2020detr}, assigning each joint to a learned query token. UnrealEgo-2~\cite{akada2024unrealego2} applies standard cross-attention for spatial reasoning, while EgoPoseFormer~\cite{yang2024epfv1} and~\cite{akada2025egorear} leverage deformable cross-attention to enhance geometric alignment via learned 2D projections. While achieving satisfactory accuracy, their disadvantages include: 1) lack of temporal modeling, 2) requiring memory-intensive multi-query mechanisms and additional models (e.g., segmentation), and 3) relying on operations that are challenging to be lowered to edge-computing devices~\cite{yang2024widthformer} (e.g., deformable attention). Our method improves upon these by using identity-conditioned single-query transformers with efficient, projection-based conditioned multi-view cross-attention and full temporal integration. 
In addition to camera-based approaches, other works estimate body pose from alternative modalities~\cite{wang2025ego4o,jiang2024egoposer,jiang2022avatarposer,du2023avatars,hong2025egolm,yi2025estimating,barquero2025sparse,starke2024categorical,parger2018human,rozumnyi2025xr} (\eg. from wearable devices).

\subsection{Datasets for egocentric motion estimation}

EgoCap~\cite{rhodin2016egocap} achieved egocentric motion capture using two fisheye cameras mounted on rigid rods extending from a helmet. Although effective for research, the setup was unsuitable for consumer devices. EgoGlass~\cite{zhao2021egoglass} later adopted a more realistic configuration with two temple-mounted cameras on glasses, but the data set was not publicly released. Given the difficulty of collecting real-world data with accurate 3D ground truth, several works turned to simulation. Mo$^2$Cap$^2$~\cite{xu2019mo2cap2} created synthetic single-camera data to predict 3D joint distances, while xR-EgoPose~\cite{tome2019xregopose} and UnrealEgo~\cite{akada2022unrealego,akada2024unrealego2} generated large-scale synthetic imagery under headset-like camera placements.  EgoRear~\cite{akada2025egorear} adds rear views to the frontal inputs, improving body coverage during occlusions. EgoBody3M~\cite{zhao2024egobody3m} represents the first large-scale real-world dataset captured directly on a VR headset using four synchronized cameras, providing millions of frames with accurate 3D ground truth. It enabled benchmarking under realistic optics and head motion, but the collection process required extensive and non-trivial labor, making large-scale expansion impractical. Other concurrent works, such as EgoHumans~\cite{khirodkar2023egohumans} and Ego4D~\cite{grauman2022ego4d}, captured naturalistic egocentric scenes but lacked dense and accurate 3D body labels. Nymeria {\cite{ma2024nymeria} and EMHI~\cite{fan2025emhi} provided large-scale real-world datasets.

\subsection{Auto-labeling}

Our auto-labeling system follows the semi-supervised learning (SSL) paradigm~\cite{chapelle2009ssl}, which aims to leverage large volumes of unlabeled data to improve model performance. Among various SSL approaches~\cite{kingma2014ssvaes,rasmus2015ladder,sajjadi2016pimodel,liu2020catgan,chen2020datasemi}, auto-labeling (\ie, pseudo-labeling) has shown its advantage because of its simplicity, scalability, and strong empirical results. The core idea is firstly training a teacher model on labeled data, then using it to generate pseudo labels for unlabeled inputs, and finally training a student model with both ground-truth and pseudo supervision~\cite{yang2022sslsurvey,kage2024sslreview}. Tremendous progress has been achieved by investigating sample scheduling~\cite{berthelot2019mixmatch,berthelot2019remixmatch,cubuk2019autoaugment}, confidence-based selection~\cite{sohn2020fixmatch,zhou2020timeconsist}, curriculum learning~\cite{cascante2021curriculum,zhang2021flexmatch}, \textit{etc}. Importantly, this strategy has been successfully applied to various computer vision tasks~\cite{tarvainen2017meanteacher,kirillov2023sam,yang2022st++,zhou2022dense,zhang2023semidetr,li2024patchrefiner}. Surprisingly, we notice that this line of research remains under-explored for egocentric motion estimation, likely due to the scarcity of large-scale egocentric datasets. The most relevant work is by EgoPW~\cite{wang2022egopw}, which introduces an in-the-wild egocentric dataset paired with external-view images to generate pseudo labels via multi-view optimization. Their model is trained using adversarial domain adaptation with features extracted from an off-the-shelf external pose estimator~\cite{xiao2018simple}. In contrast, our system relies solely on egocentric inputs without any auxiliary outside-in views, multi-view optimization, or external networks, making it more scalable and applicable to real-world deployment scenarios.  

\section{Method}
\label{sec:method}

\begin{figure*}
    \centering
    \includegraphics[width=0.99\linewidth]{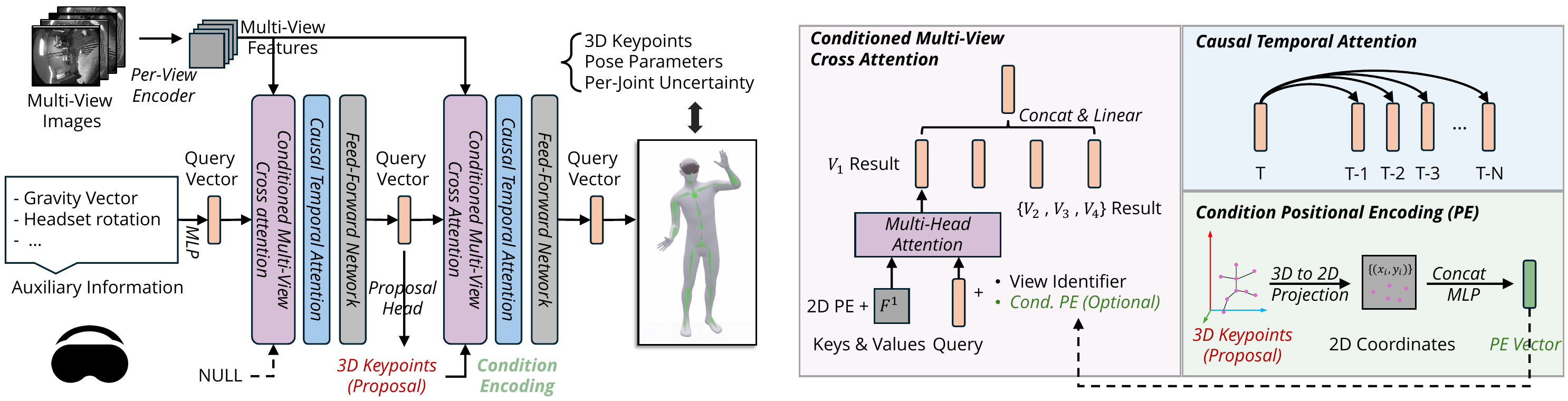}
    \vspace{-8pt}
    \caption{\textbf{Architecture overview (left).} We stack two transformer decoders for coarse-to-fine pose estimation. A single holistic query, initialized from auxiliary metadata, attends to multi-view features and historic information to estimate 3D keypoints, pose parameters, and per-joint uncertainty in an end-to-end differentiable architecture. \textbf{Illustration of the two core attention modules (right)}. Causal temporal attention enables each frame to attend to its temporal history. Conditioned multi-view cross attention incorporates both view identity and optional 2D keypoint projections of pose proposal to guide spatial feature aggregation across views.}
    \label{fig:framework}
    \vspace{-12pt}
\end{figure*}

We define real-time egocentric motion capturing as follows: given the synchronized input stream $\inputset_t=\bigl\{\{\inputimage^v_t\}_{v=1}^V, \inputheadset_t\bigr\}$ at timestamp $t$, where $\inputimage^v$ denotes the calibrated image captured by the $v$-th camera, and $\inputheadset$ denotes the 6DoF headset pose provided by the headset system~\cite{zhao2024egobody3m}, the goal is to reconstruct the user's pose in the world space. Here we consider two types of \textit{pose representations}.
The first type is the keypoint pose $\outputpose_t \in \R^{\jointnum\times 3}$ where $\jointnum$ denotes the number of body keypoints; the second type of poses are joint rotations and body scales $\theta_t \in \R^{\rotatenum}$ where $\rotatenum$ is predefined by the body model~\cite{momentum}. Note that we can also compute keypoints from joint rotations through \textit{forward kinematics}.

\subsection{Prelimilary}
\label{sec:method:epfv1}
Our model architecture~is built upon the state-of-the-art EgoPoseFormer model (shorten as EPFv1). Below, we will provide a brief recap of the baseline architecture.

\noindent \textbf{Overview.} Given the multi-view images $\{\inputimage^v\}_{v=1}^V$\footnote{EPFv1 is a single-frame model so we omit the temporal footnote here.}, EPFv1 first uses an image encoder to extract the multi-view features $\{\mathbf{F}^v\}_{v=1}^V$, then it computes the \textit{pose proposal} $\tilde{p} \in \R^{\jointnum\times 3}$, \ie, the roughly estimated 3D positions of body keypoints, from the pooled image features using a 2-layer MLP:
\begin{align}
    \tilde{\mathbf{p}} = \mathrm{MLP}\Bigl(\mathrm{Concat}_{\{v\}}\bigl(\mathrm{AvgPool}(\mathbf{F}_v)\bigr)\Bigr).
\end{align}

Then, EPFv1 encodes each body keypoints into a \textit{Joint Query Token (JQT)}, which servers as the query token to interact with the image features and other body joints. Specifically, for the $j$-th keypoint, its JQT is computed by:
\begin{align}
    \tilde{\mathbf{q}}_j= \mathrm{MLP}_{\mathrm{JQT}}\bigl(\eta_j, \tilde{x}_j, \tilde{y}_j, \tilde{z}_j\bigr),
\end{align}
where $\eta_j$ is an scalar keypoint identifier. Then, the JQTs are input into the \textit{Pose Refinement transformer (PRFormer)} to extract useful information. Finally, a 3D refinement offset is predicted for each keypoint, and its final 3D position is computed by adding the offset to the proposal:
\begin{align}
    \mathbf{q} = \mathrm{RefineTransformer}&\bigl(\tilde{\mathbf{q}}, \{\mathbf{F}_v\}_{v=1}^V\bigr), \\
    \Delta \mathbf{p}_j = \mathrm{MLP}_{\mathrm{offset}}\bigl(\mathbf{q}_j\bigr),~~&~\mathbf{p} = \tilde{\mathbf{p}} + \Delta \mathbf{p}
\end{align}

\noindent \textbf{Deformable stereo attention.} 
EPFv1 uses a special attention mechanism called \textit{Deformable Stereo Attention} inside each layer of its PRFormer, which serves as the key factor to EPFv1's good accuracy over previous works. Specifically, each 3D proposal $\tilde{p_j}$ is first projected to each image plane to compute its 2D coordinates $\{\mathbf{o}_j^v\}_{v=1}^V$. Those 2D keypoints then serve as the anchor point, based on which deformable attention~\cite{zhu2020deformable} is used to extract information from each image features. Finally a linear layer is used to fuse the multi-view information:
\begin{align}
    \mathbf{q}_j = \textrm{Linear}\Bigl(\textrm{Concat}_{\{v\}}\bigl(\textrm{DeformAttn}(\mathbf{q}_j, \mathbf{o}_j^v, \mathbf{F}_v)\bigr)\Bigr).
\end{align}
Through this process, the multi-view stereo features are effectively exploited by the attention mechanism, thus the model can achieve accurate 3D keypoints localization. 

\noindent \textbf{Discussion.}
Although EPFv1~\cite{yang2024epfv1} and the follow-up methods (e.g., EgoRear~\cite{akada2025egorear}) achieved good accuracy on open-sourced benchmarks~\cite{akada2022unrealego,wang2023scene}, several remaining limitations still hinder their usage on real XR devices~\cite{zhao2024egobody3m}. Specifically: 1) Their prediction lack temporal smoothness because they do not have any temporal modeling~\cite{zhao2024egobody3m}, 2) Their one-to-one correspondence between body-keypoints and joint query tokens limit the model efficiency, \ie, the model's computation cost grows linearly with the number of predicted keypoints; 
3) This one-to-one correspondence also limits the model's flexibility to be used for different body representations,
\eg, pose parameters~\cite{momentum, pavlakos2019expressive}, because they are not able to project those parameters into 2D planes. 4) The models are not fully end-to-end trainable, for example, there is no gradient flow from the EPFv1's pose refinement transformer to the pose proposal network, thereby limiting the model's capacity. In the next section, we demonstrate how \ourmethod~overcomes those challenges and enables accurate body motion estimation for real XR-devices.

\subsection{\ourmethodtitle}

As shown in Fig.~\ref{fig:framework}, our model has an encoder-decoder structure. At timestamp $t$, the encoder separately encodes the images $\{\inputimage^v_t\}_{v=1}^V$ into feature maps $\{\featuremap_t^v\}_{v=1}^V$, and headset pose $\inputheadset_t$ and other optional auxiliary information into a holistic pose query token $\mathbf{q}_t \in \R^{C}$. Our pose decoder share a similar design to EPFv1, which also has a pose proposal stage and a pose refinement stage. However, unlike EPFv1's MLP + Transformer design, here we stack two architecturally identical transformer decoders for both stages, making the architecture even simpler. After each transformer, we use MLP heads to compute the task results, \eg, keypoints, pose parameters and uncertainties.

\subsubsection{Pose query token} 

As discussed in Sec.~\ref{sec:method:epfv1}, EPFv1's 1:1 correspondence between body keypoints and transformer queries limits its computation efficiency and flexibility. Therefore, in \ourmethod, we turn to using a single holistic pose query token to serve as an information aggregator, and all task predictions are computed from it, decoupling the model efficiency from the body representation and task definition. Moreover, this design also allows us to seamlessly inject \textit{auxiliary information} into the model, \eg, headset poses, which can be incorporated simply by encoding them with an MLP:
\begin{align}
    \mathbf{q}_t = \mathrm{MLP}_{\mathrm{query}}(\inputheadset_t).
\end{align}
If no auxiliary information is accessible, we modify $\mathbf{q}_t$ to be a learnable and randomly initialized embedding. 

\subsubsection{Pose decoding transformer}

Shown in Fig.~\ref{fig:framework}, each pose decoding transformer layer has three modules: 1) a conditioned multi-view cross-attention, 2) a causal temporal attention, and 3) a feedforward network (FFN). As FFN is a standard block in transformers, we will only explain the attention modules. 

\noindent \textbf{Conditioned multi-view cross-attention.}
\ourmethod~uses standard \textit{`query to image'} cross-attention~\cite{kirillov2023sam, carion2020end} instead of deformable attention used by EPFv1, whose deployment on edge-computing devices is challenging~\cite{yang2024widthformer}. To make our cross-attention handle multi-view images, we simply do multi-head attention on each images independently, and fuse the multi-view information using the output linear layer. Formally, given the pose query token $\mathbf{q}_t$ and image features $\{\featuremap_t^v\}_{v=1}^V$, the result token $\mathbf{q}'_t$ is computed by:
\begin{align}
\resizebox{.9\linewidth}{!}{$
    \mathbf{q}'_t = \mathrm{Linear}\Bigl(\mathrm{Concat}_{\{v\}}\bigl(\mathrm{\hat{MHA}(\mathbf{q}_t+\mathbf{\sigma}_t^v, \featuremap_t^v+\mathbf{\Psi}, \featuremap_t^v)}\bigr)\Bigr).
$}
\end{align}
Here, $\hat{\mathrm{MHA}}(q,k,v)$ is the multi-head attention but without the output linear layer~\cite{vaswani2017attisallyouneed}, $\mathbf{\Psi}$ is the 2D image positional encoding~\cite{dosovitskiy2020vit}, and $\mathbf{\sigma}_t^v$ is the query conditional embedding, which is used to achieve \textit{conditioned attention}. Specifically, we consider two types of conditions: 1) camera identities in both pose proposal stage and pose refinement stage, 2) projected 2D keypoints only in the pose refinement stage:
\begin{align}
\resizebox{.9\linewidth}{!}{$
\sigma_t^v = 
\begin{cases}
    \mathbf{\xi}^v ~~~~~~~~~~~~~~~~~~~~~~~~~~~~~~~~~&\text{Pose proposal}\\
    \mathbf{\xi}^v + \mathrm{MLP}\bigl(\mathrm{Concat}_{\{j\}}(\mathbf{o}_{t,j}^v)\bigr)~&\text{Pose refinement}
\end{cases}
$}
\end{align}
where $\xi^v$ is a learnable camera embedding for the $v$-th camera, and $\mathbf{o}_{t,j}^v$ is the 2D position of the $j$-th keypoint on the $v$-th image, computed by projecting 3D proposal keypoints onto the 2D plane. Here, by encoding the concatenated 2D keypoints into a latent conditional vector, we achieve a similar effect to EPFv1's deformable stereo attention but in a hardware-friendly way, in which our model also exploit the stereo information to predict 3D pose accurately.

\noindent \textbf{Causal temporal attention.}
To ensure \ourmethod~can output smooth body poses, we use a causal temporal attention to enable the pose query token to attend to its history. This temporal attention also enables the model to infer plausible body poses from temporal clues when certain body-parts are invisible, \eg, when they leave the camera FoV. Specifically, at timestamp $t$, the query token $\mathbf{q}_t$ attends to the past query tokens in a pre-defined window size $w$:
\begin{align}
\resizebox{.9\linewidth}{!}{$
    \mathbf{q}'_t = \mathrm{MHA}\Bigl(\mathrm{RoPE}(\mathbf{q}_t), \mathrm{RoPE}(\{\mathbf{q}_k\}_{k=t-w}^{t}), \{\mathbf{q}_k\}_{k=t-w}^{t-1}\Bigr).
$}
\end{align}
Here $\mathrm{RoPE()}$ is the transformation of rotational positional encoding~\cite{su2024roformer}. Similar to the auto-regressive language models~\cite{achiam2023gpt,dubey2024llama}, during training the model receives a sequence of images, thus the temporal attention becomes a standard self-attention, and we use attention masks to mask out the invalid attention weights, \eg, past to future. During inference, we use KV-Cache~\cite{dubey2024llama} to store the historical keys and values, saving computational resources.

\begin{figure}
    \centering
    \includegraphics[width=\linewidth]{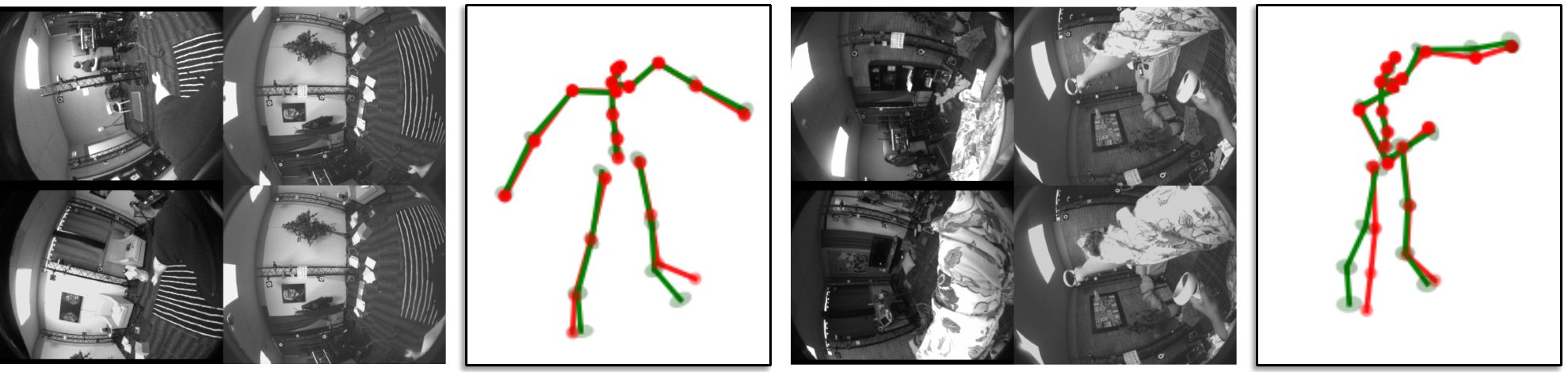}
    \vspace{-15pt}
    \caption{\small \textbf{Per-keypoint uncertainty predicted by \ourmethod.} Larger ellipse extent and higher transparency indicate higher predicted uncertainty. Prediction is in \textcolor{ForestGreen}{green} whereas GT is in \textcolor{BrickRed}{red}.}
    \label{fig:uncertainty}
    \vspace{-15pt}
\end{figure}

\subsubsection{Task heads}
After each pose decoder stage, we feed the pose query token into parallel MLP heads to compute task predictions. 

\noindent \textbf{Pose proposal.}
For the pose proposal stage, we follow \mbox{EgoBody3M}~\cite{zhao2024egobody3m} to directly regress the 3D body keypoints in the headset space, which we found is good enough for the pose refinement transformer to be conditioned on. 

\noindent \textbf{Pose refinement.}
The pose refinement decoder computes the final body pose in the world coordinate space, for which we first predict body poses in the headset space and convert them to world space using the 6-DoF headset poses~\cite{zhao2024egobody3m}. As a flexible architecture, \ourmethod~supports predicting both 3D keypoints and pose parameters~\cite{momentum}. For 3D keypoints, we follow EPFv1~\cite{yang2024epfv1} to predict a refinement offset w.r.t. each pose proposal keypoint, then the final 3D keypoints in the headset space are computed by adding the offset to the proposal 3D keypoints. When using pose parameters, the model make three predictions: 1) joint rotations, 2) body scales, and 3) 6-DoF transformation from head to headset, for which the model predict 3D translations and 6D rotations~\cite{zhou2019continuity}. Then, we run \textit{Forward Kinematics} to compute the 3D joint positions and rotations in the world space.

\noindent \textbf{Per-keypoint uncertainty.} Apart from the body poses, \ourmethod~also supports predicting per-joint uncertainty in the headset pace, which improves the model's accuracy and helps with the auto-labeling system (Sec.~\ref{sec:als}). Specifically, we extend the LUVLi loss~\cite{kumar2020luvli} to 3D space. For each keypoint, the model predicts a 6D uncertainty $\uncertainty \in\R ^6$, representing the lower-triangular Cholesky factor $\cholesky$ of the covariance $\covariance = \cholesky\cholesky^\text{T}$. The likelihood loss is formulated as
\begin{align}
    &\residual_j= \hat{\mathbf{p}}_j - \mathbf{p}_j,~~~\mathbf{m} = \residual^\text{T}\covariance^{-1}\residual, \\
    \mathcal{L}_\mathrm{{tNLL}}=&\frac{(\studenttv+\studenttd)}{2}\log(1+\frac{\mathbf{m}}{\studenttv}) + \frac{1}{2}\log{|\covariance|}
\end{align}
where $\hat{\mathbf{p}}_j$ and $\mathbf{p}_j$ are the ground-truth and predicted 3D positions, and dimensionality $\studenttd=3$. The first term is a scaled Mahalanobis distance, whereas the second term serves as a regularization that ensures that the uncertainty distribution does not get too large. Different from LUVLi ~\cite{kumar2020luvli} we consider the student-t distribution where $\studenttv$ controls the tail heaviness. It has the same properties as the Laplacian distribution used in \cite{kumar2020luvli}, but is smoother at the origin and more heavy‑tailed for large residuals. As shown in~\cref{fig:uncertainty}, \ourmethod\ produces meaningful uncertainty estimates, particularly for invisible keypoints, e.g., feets and legs.

\noindent \textbf{Losses}: \ourmethod outputs the pose proposal $\mathbf{P}_p$, the refined pose $\mathbf{P}_r$ (from keypoint refinement or forward kinematics), and the per-keypoint uncertainty $\covariance$ for the refined pose. Given the ground-truth pose $\mathbf{\hat{P}}$, the overall training objective is defined as a weighted sum of several loss terms:
{\small
\begin{align}\label{eq:total_loss}
    \mathcal{L} = & \lambda_{\text{pos}}w_d\mathcal{L}_{\text{mse}}(\mathbf{P}_{r}, \mathbf{\hat{P}}) + \lambda_{\text{pos}}(1 - w_d)\mathcal{L}_{\text{tNLL}}(\mathbf{P}_{r}, \mathbf{\hat{P}}, \covariance) \nonumber \\
    &  + \lambda_{\text{pos}}\mathcal{L}_{\text{mse}}(\mathbf{P}_{p}, \mathbf{\hat{P}}) + \lambda_{\text{jerk}}\mathcal{L}_{\text{jerk}}(\mathbf{P}_{r}) + \lambda_{\text{jerk}}\mathcal{L}_{\text{jerk}}(\mathbf{P}_{p})
\end{align}
}where we follow EgoBody3M~\cite{zhao2024egobody3m} to adopt the constant pose jerk loss $\mathcal{L}_{\text{jerk}}$ to encourage temporally smooth predictions. In practice, we set $\lambda_{\text{pos}}=1$ and $\lambda_{\text{jerk}}=0.8$, respectively. During training, we combine the MSE and uncertainty Log-Likelihood losses with a dynamic weight $w_d$ that follows a cosine schedule. It gradually increases the contribution of the Log-Likelihood term while decreasing the MSE term as training progresses. In practice, we start with a detached uncertainty head and gradually reduce the degree of detach using soft detach to stabilize training.

\begin{figure}
    \centering
    \includegraphics[width=0.98\linewidth]{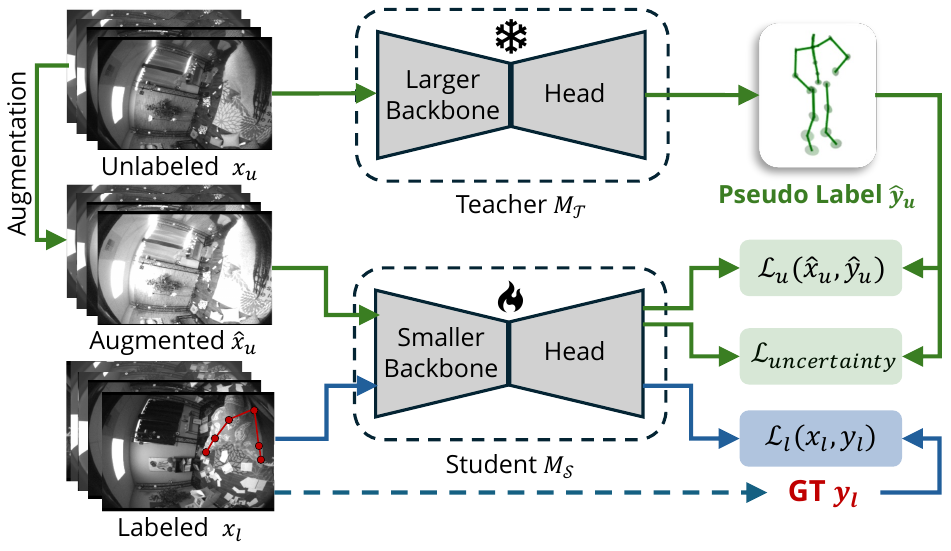}
    \vspace{-6pt}
    \caption{\textbf{Overview of the mixture training in auto-labeling system.} We adopt a stronger teacher model for pesudo labeling and apply an uncertainty distillation loss to facilitate the knowledge transfer. The teacher model is pre-trained with the labeled dataset $\mathcal{D}_{l}=\{(x_l, y_l)\}$ before this stage.}
    \label{fig:ssl}
    \vspace{-12pt}
\end{figure}

\subsection{Auto-labeling system} 
\label{sec:als}
Collecting large-scale, labeled egocentric motion data remains costly and labor-intensive, especially under real-world capture conditions~\cite{azam2024survey,li2025challenges}. Therefore, we adopt an auto-labeling system that  fully utilize the data to improve the model performance through semi-supervised learning (SSL)~\cite{chapelle2009ssl}. This design enables us to benefit from high-capacity models and large training corpora without requiring additional human annotations.

Formally, let $\mathcal{D}_{l} = \{(x_l, y_l)\}$ denote the labeled source dataset and $\mathcal{D}_{u} = \{(x_u)\}$ the unlabeled dataset. The system operates in two stages: (1) training a high-quality teacher model $\mathcal{M}_\mathcal{T}$ on $\mathcal{D}_{l}$ using vision foundation models~\cite{simeoni2025dinov3,radford2021clip}, and (2) using it to supervise a student model $\mathcal{M}_\mathcal{S}$ with a combination of labeled and pseudo-labeled samples.

As shown in Fig.~\ref{fig:ssl}, each batch during the SSL training consists of labeled data $(x_l, y_l)$ and unlabeled data $x_u$. We generate a strongly augmented version $\hat{x}_u$ of each unlabeled input. The teacher model  receives the original input $x_u$ and predicts pseudo labels $\hat{y}_u = \mathcal{M}_\mathcal{T}(x_u)$. These pseudo labels are then used to supervise the student model $\mathcal{M}_S$, which processes the corresponding augmented input $\hat{x}_u$. This asymmetric design, following previous work~\cite{tarvainen2017meanteacher,zhou2022dense,li2022stmono3d}, ensures that the teacher generates stable targets while the student learns to generalize under appearance variation. 

In addition to pose supervision, we introduce an auxiliary uncertainty distillation loss between the teacher and student predictions. Specifically, we treat each model’s uncertainty output as a soft confidence map over keypoints and minimize a MSE loss between their predicted uncertainties:
\begin{equation}
\mathcal{L}_{\text{uncertainty}} = || \uncertainty_T - \uncertainty_S ||,
\end{equation}
where $\uncertainty_\mathcal{T}$ and $\uncertainty_\mathcal{S}$ denote the predicted per-joint uncertainty vectors from the teacher and student models, respectively. This regularization encourages the student to mimic not only the pose estimates but also the teacher’s confidence structure, facilitating effective knowledge transfer from teacher to student. The total training objective for the student model can be formulated as:
\begin{equation}
\mathcal{L}_{\text{semi}} = \mathcal{L}_{l}(x_l, y_l) + \lambda_1 \cdot \mathcal{L}_{u}(\hat{x}_u, \hat{y}_u) + \lambda_2 \cdot \mathcal{L}_{\text{uncertainty}},
\end{equation}
where $\lambda_1=0.5$ and $\lambda_2=0.1$ balance the contributions of different loss terms. Both $\mathcal{L}_{l}$ and $\mathcal{L}_{u}$ follow the same multi-term pose loss formulation defined in Eq.~\ref{eq:total_loss}, applied respectively on labeled and pseudo-labeled samples.

\section{Experiments}
\label{sec:exps}
\vspace{-5pt}

\subsection{Experimental setup}
\vspace{-2pt}

\begin{table*}[t!]
    \caption{\textbf{Quantitative comparison on Egobody3M benchmark.} The percentage in parentheses denotes how much worse each baseline is compared to our method. $^{\dag}$: indicates our reproduced results. $^{\ddag}$: presents results reported in the Egobody3M paper with missing metric numbers filled in with $-$. ALS is short for our Auto-Labeling System.} 
    \label{tab:egobody3m}
    \vspace{-2pt}
    \centering
    \scalebox{0.69}{
    \begin{tabular}{L{3.2cm}|C{2.3cm}*{2}{|L{2.7cm}}*{4}{|L{2.7cm}}}
        \toprule
        \multirow{2}{*}{Method} & \multirow{2}{*}{Reference} & \multicolumn{2}{c|}{Overall}  & \makecell[c]{Overall Wrists} & \makecell[c]{Overall Shoulders} & \makecell[c]{Overall Legs} & \makecell[c]{Overall Feet} \\
        \cmidrule(l){3-8}
        &  & \makecell[c]{MPJPE} & \makecell[c]{MPJVE} & \makecell[c]{MPJPE} & \makecell[c]{MPJPE} & \makecell[c]{MPJPE} & \makecell[c]{MPJPE} \\
        \midrule
        UnrealEgo$^{\ddag}$~\cite{akada2022unrealego} & ECCV2022 & ~~7.41 $(\textcolor{BrickRed}{+84.3\%})$ & ~~1.27 $(\textcolor{BrickRed}{+202.4\%})$ & \makecell[c]{$-$} & \makecell[c]{$-$} & \makecell[c]{$-$} & \makecell[c]{$-$} \\
        Egobody3M$^{\dag}$~\cite{zhao2024egobody3m} & ECCV2024 & ~~5.18 $(\textcolor{BrickRed}{+28.9\%})$ & ~~0.54 $(\textcolor{BrickRed}{+28.6\%})$ & ~~6.14 $(\textcolor{BrickRed}{+23.0\%})$ & ~~2.80 $(\textcolor{BrickRed}{+19.9\%})$ & ~~8.40 $(\textcolor{BrickRed}{+26.1\%})$ & ~~10.25 $(\textcolor{BrickRed}{+17.9\%})$ \\ 
        EgoPoseFormer$^{\dag}$~\cite{yang2024epfv1} &  ECCV2024 & ~~4.75 $(\textcolor{BrickRed}{+18.1\%})$ & ~~0.87 $(\textcolor{BrickRed}{+107.1\%})$ & ~~6.01 $(\textcolor{BrickRed}{+20.4\%})$ & ~~2.72 $(\textcolor{BrickRed}{+16.5\%})$ & ~~7.95 $(\textcolor{BrickRed}{+19.4\%})$ & ~~10.16 $(\textcolor{BrickRed}{+17.0\%})$ \\
        \midrule
        \ourmethod w/o ALS &  - & ~~4.17 $(\textcolor{BrickRed}{+3.2\%})$ & ~~0.42 $(\textcolor{gray}{+0.0\%})$ & ~~5.74 $(\textcolor{BrickRed}{+15.3\%})$ & ~~2.38 $(\textcolor{BrickRed}{+2.1\%})$ & ~~6.91 $(\textcolor{BrickRed}{+3.8\%})$ & ~~~~9.11 $(\textcolor{BrickRed}{+4.8\%})$ \\ 
        \ourmethod with ALS & - & ~~\makecell[l]{\textbf{4.02}} & ~~\makecell[l]{\textbf{0.42}} & ~~\makecell[l]{\textbf{4.99}} & ~~\makecell[l]{\textbf{2.33}}  & ~~\makecell[l]{\textbf{6.66}}  & ~~\makecell[l]{~~\textbf{8.69}}  \\
        \bottomrule
    \end{tabular}
    }
    
    \vspace{-10pt}
\end{table*}

\noindent\textbf{Implementation details:} We adopt a single-view encoder to extract per-frame image features. Following EPFv1~\cite{yang2024epfv1}, the encoder is first pre-trained on a 2D joint heatmap prediction task to enhance spatial awareness. For our teacher models, we initialize our ViT-based encoder with DINOv3~\cite{simeoni2025dinov3} weights to improve representation quality. For a fair comparison with prior approaches, we implement ResNet-18~\cite{he2016resnet} as the default backbone for benchmarking. In experiments, we also consider a smaller model MobileNetv4-S~\cite{qin2024mobilenetv4} and larger variant ResNet-50~\cite{he2016resnet}. All models are trained with a batch size of 16 for 40k optimization steps. We present more details in our \textbf{supplementary materials}, including detailed speed benchmark and how we build the auto labeling system.

\noindent\textbf{Datasets:} We use Egobody3M~\cite{zhao2024egobody3m} to benchmark our proposed method. It contains 3.4M real-domain frames captured by 4 synchronized monochrome global-shutter cameras. We use the standard training set with 2.4M frames to train our model and evaluate the model using the test set with 426k frames. %
In addition, as EgoBody3M only have keypoints annotations, we compute all model parameters by running \textit{inverse kinematics} using the momentum tool~\cite{momentum}. We report the Mean Per Joint Position Error (MPJPE) and Mean Per-Joint Velocity Error (MPJVE) as evaluation metrics following the official paper. To show the effectiveness of our auto-labeling system, we collect a large in-the-wild (ITW) private dataset without ground-truth labels, using the same hardware settings as EgoBody3M~\cite{zhao2024egobody3m}. This dataset, called EGO-ITW-70M, contains 70M frames of images with a diverse environment and human motions. 
Finally, to assess real-world generalization, we run our model on the in-the-wild XR-MBT dataset~\cite{rozumnyi2025xr} and show qualitative results in Fig.~\ref{fig:teaser}.

\begin{figure*}
    \centering
    \includegraphics[width=0.99\textwidth]{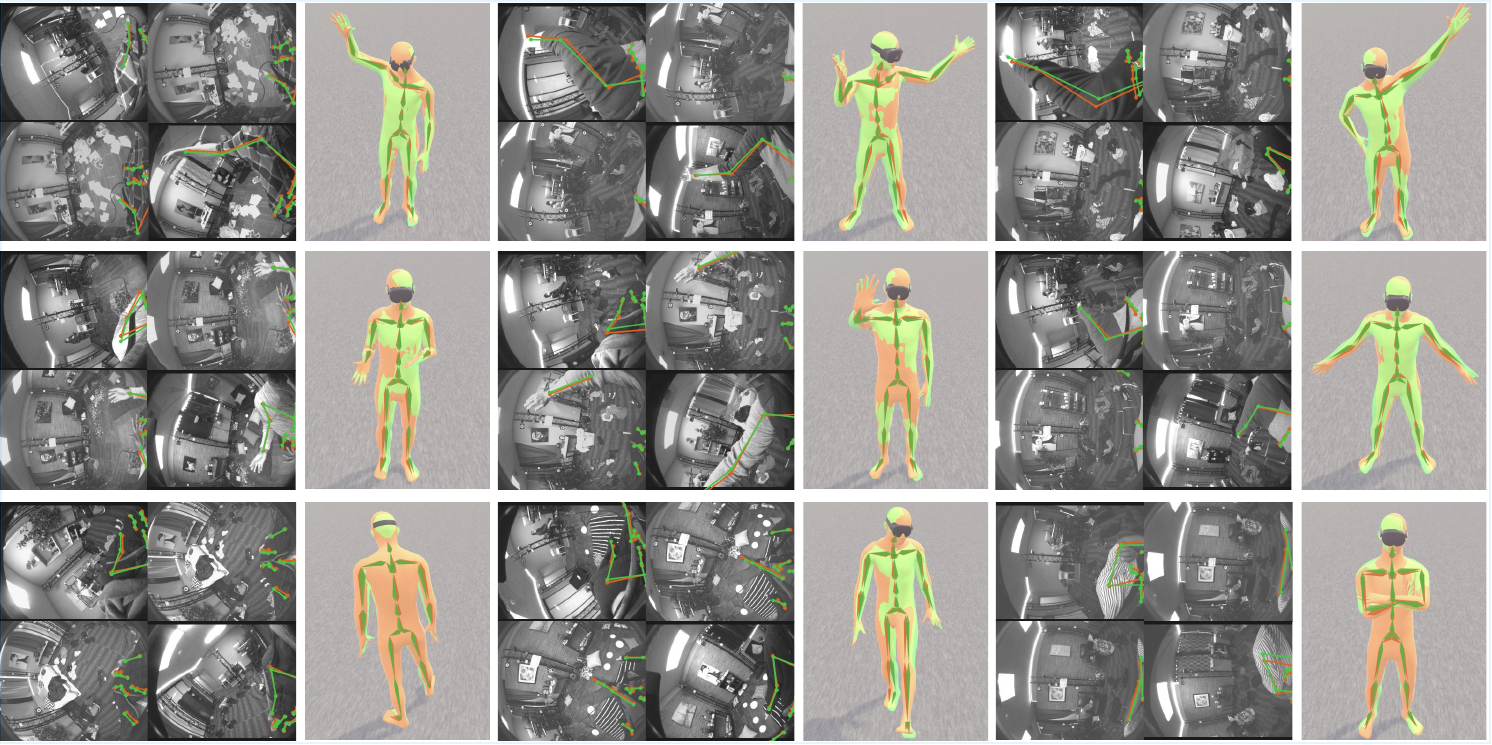}
    \vspace{-6pt}
    \caption{\textbf{Qualitative results on Egobody3M.} Predictions are colored in \textcolor{ForestGreen}{green} and ground-truths are colored in \textcolor{orange}{orange}.}
    \label{fig:qualitative}
    \vspace{-10pt}
\end{figure*}

\vspace{-2pt}
\subsection{Comparison with the state of the art}
\vspace{-2pt}
In \cref{tab:egobody3m}, we compare \ourmethod with prior SoTA approaches on the EgoBody3M benchmark~\cite{zhao2024egobody3m}. We reproduce the results of EgoBody3M and EgoPoseFormer~\cite{yang2024epfv1}, while the result for UnrealEgo is taken from the original EgoBody3M paper. Our method outperforms all competing methods across all metrics with a clear margin. Specifically, \ourmethod achieves a mean per-joint position error (MPJPE) of 4.02cm, improving upon EgoBody3M and EgoPoseFormer by 22.4\% and 15.4\%, respectively. We also report the MPJPE for wrist joints, which are challenging due to frequent occlusions and rapid motion. Our method achieves a wrist MPJPE of 4.99cm, significantly outperforming prior works by more than 15\%. In terms of temporal stability, our model achieves an MPJVE  reduction of 22.2\% compared to EgoBody3M and 51.7\% compared to EgoPoseFormer. This highlights the importance of temporal modeling for egocentric pose tracking and demonstrates the effectiveness of our causal temporal attention design. Fig.~\ref{fig:qualitative} showcases qualitative examples of \ourmethod in challenging scenarios with occlusion, fast motion, and body truncation. These visualizations confirm the robustness and precision of our approach for egocentric full-body pose estimation.

\subsection{Ablation study and discussion}
\label{subsec:ablation}
We conduct several ablation experiments to analyze the contribution of each component in \ourmethod. Unless otherwise stated, we adopt ResNet-18 as backbone and use the Egobody3M dataset by default.

\begin{table}[t!]
    \caption{\textbf{Ablation study.} We remove or modify each module to assess its contribution. These results demonstrate that each component yields a measurable improvement in accuracy and our auto-labeling system further boosts the model performance.}
    \label{tab:main_ablation}
    \centering
    \scalebox{0.70}{
    \begin{tabular}{C{0.3cm}|L{4.2cm}|C{1.5cm}|C{1.5cm}|C{1.5cm}}
        \toprule
        \# & Variants & Overall & \makecell[c]{Wrists} & \makecell[c]{Legs} \\
        \midrule
        \ding{172} & use \textit{kpt} head & 4.35  & 6.02 & 7.17 \\ 
        \ding{173} & w/o temporal attention  & 4.35  & 6.04 & 7.21 \\ 
        \ding{174} & w/o projection condition  & 4.30  & 5.96 & 7.15 \\ 
        \ding{175} & w/o auxiliary information  & 4.39  & 5.98 & 7.34 \\ 
        \ding{176} & w/o uncertainty & 4.25  & 5.83 & 7.00 \\ 
        \midrule
        \ding{177} & \ourmethod w/o ALS & 4.17 & 5.74 & 6.91 \\
         \midrule
        \ding{178} & + ALS (w/o $\mathcal{L}_{uncertainty}$) & 4.08 & 5.07 & 6.74 \\
        $\bigstar$ & + ALS + $\mathcal{L}_{uncertainty}$ & \textbf{4.02} & \textbf{4.99} & \textbf{6.66} \\
        \bottomrule
    \end{tabular}
    }
    \vspace{-5pt}
\end{table}

\noindent\textbf{Architecture design:} 
Table~\ref{tab:main_ablation} presents ablations on the main architectural components of \ourmethod. We ablate each module by removing it from the comprehensive architecture (\ding{177}) alternately. Specifically, we examine the following settings: \ding{172} direct keypoint prediction through a \textit{kpt} head (\ie, use \textit{kpt} head) instead of leveraging joint angle prediction in combination with \textit{Forward Kinematics}, \ding{173} removing the temporal attention module (\ie, w/o temporal attention), \ding{174} removing the projected 2D keypoints as condition for the multi-view cross-attention in the pose refinement stage (\ie, w/o projection condition), \ding{175} replacing the query encoding auxiliary information with a standard learnable one (\ie, w/o auxiliary information), \ding{176} removing the uncertainty head (\ie, w/o uncertainty). 

Each component yields a measurable improvement in accuracy. Specifically, \ding{172} highlights the effectiveness of predicting full body pose parameters over directly regressing joint positions. Using forward kinematics (FK) allows the model to provide more structured and physically plausible pose predictions. Removing temporal attention (\ding{173}) reduces the model to a single-frame variant, resulting in a substantial accuracy drop, underscoring the importance of temporal modeling for accurate egocentric 3D motion estimation.
Ablating the projection-conditioned cross-attention in the refinement transformer (\ding{174}) causes the module to degenerate into a generic attention mechanism, as in the proposal stage. The model then lacks spatial anchoring cues and loses the ability to refine joint localization effectively.
The auxiliary metadata used to initialize the body query (\ding{175}) provides critical context for egocentric perception. Replacing this with a generic learnable token significantly impairs performance, indicating that this conditioning is key for learning pose-aware priors.
Finally, \ding{176} demonstrates that incorporating per-joint uncertainty helps the model focus on reliable joints during training, improving robustness and overall accuracy.

\begin{figure}
    \centering
    \includegraphics[width=0.69\linewidth]{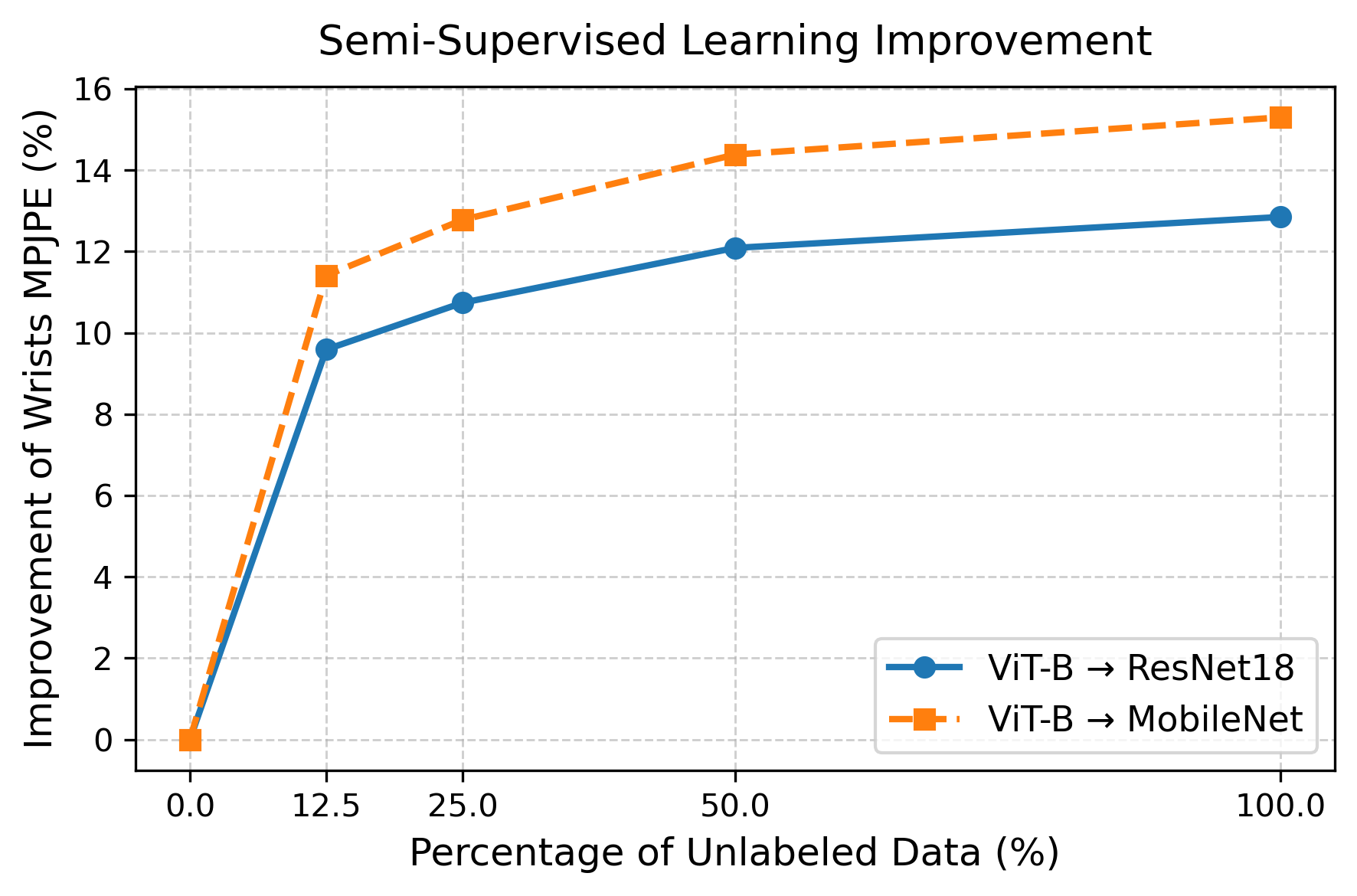}
    \vspace{-10pt}
    \caption{\textbf{ALS effectiveness in in-domain scaling.} As more unlabeled data is used, both students achieve improved accuracy. Notably, the MobileNetv4-S-based model benefits more proportionally from ALS despite having lower model capacity, indicating the pipeline’s suitability for lightweight deployment models.}
    \label{fig:ssl-performance}
    \vspace{-15pt}
\end{figure}

\noindent\textbf{Auto-labeling system (ALS):} We evaluate the proposed ALS under the \textit{data scaling} setting: we use ITW-70M as the unlabeled dataset, and adopt the Egobody3M as the labeled dataset for mixture training. Fig.~\ref{fig:ssl-performance} presents that ALS improves model performance as more unlabeled data is incorporated. We further examine the impact of different student models under a fixed DINOv3-L teacher. Interestingly, MobileNetV4-S, despite being smaller than ResNet-18, benefits more proportionally from pseudo-label supervision, suggesting that ALS is particularly effective for lightweight models in low-capacity regimes. As shown in Tab.~\ref{tab:main_ablation}, ALS (\ding{178}) yields an 11.7\% reduction in wrist MPJPE. Adding the uncertainty distillation further enhances performance slightly, indicating its complementary benefit.

\section{Conclusion}
\label{sec:conclusion}
\vspace{-5pt}

We presented EgoPoseFormer v2 (EPFv2), a method for egocentric 3D motion estimation that combines an advanced transformer architecture with a scalable auto-labeling system. We introduced a full end-to-end spatiotemporal transformer with identity-conditioned queries and efficient 3D-aware refinement. Through the uncertainty-aware auto-labeling system, \ourmethod effectively leverages unlabeled data to scale-up its ability. Experiments on the EgoBody3M benchmark demonstrate SoTA accuracy, validating the power of combining architectural advances with scalable data-driven learning. \ourmethod offers an extensible foundation for real-world AR/VR applications.

{
    \small
    \bibliographystyle{ieeenat_fullname}
    \bibliography{main}
}

\appendix
\clearpage
\setcounter{page}{1}

\twocolumn[{%
\renewcommand\twocolumn[1][]{#1}%
\maketitlesupplementary

\begin{center}
    \centering
    \captionsetup{type=figure}
    \includegraphics[width=\textwidth]{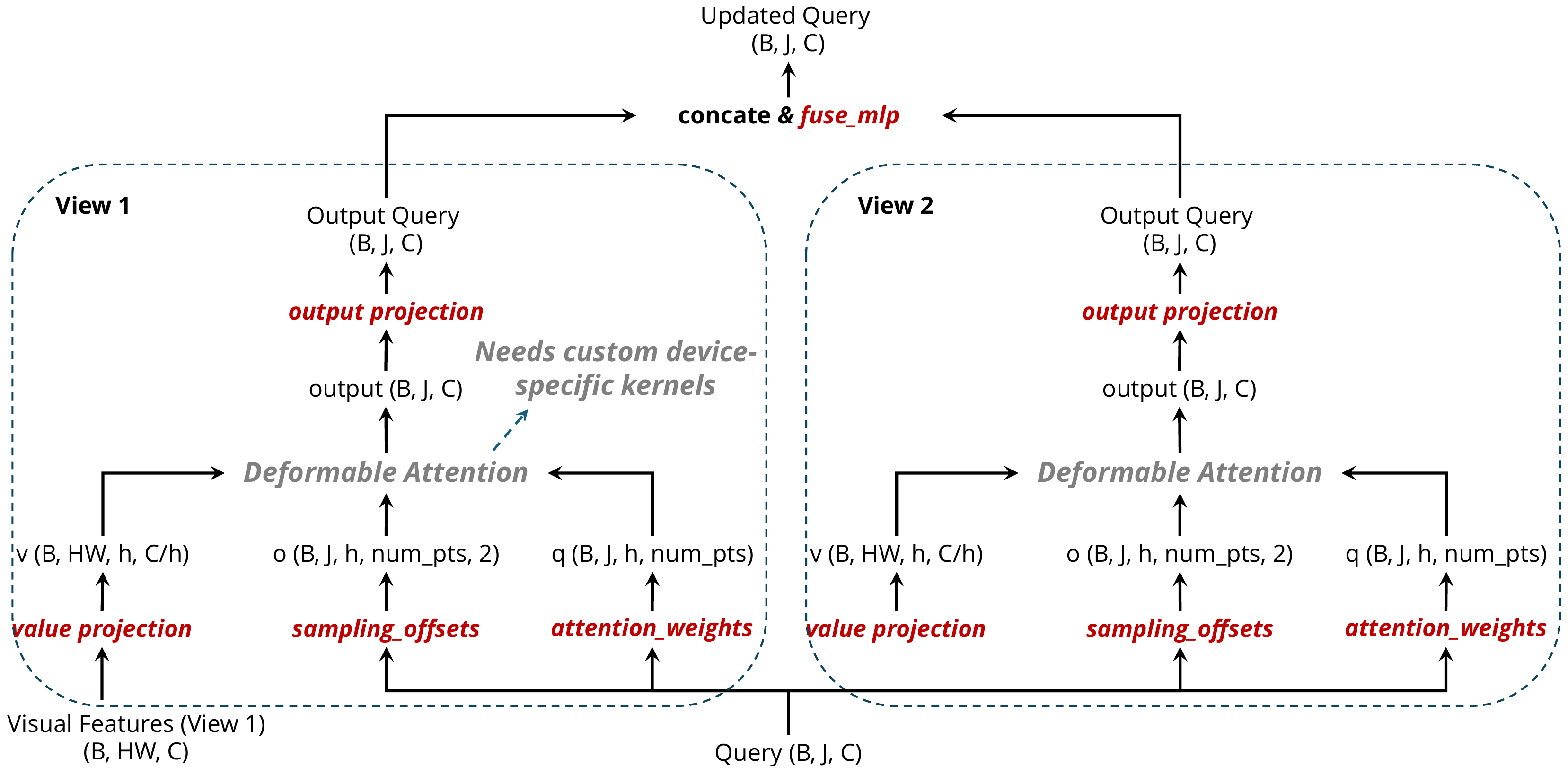}
    \vspace{-15pt}
    \captionof{figure}{Deformable stereo attention module used in EPFv1~\cite{yang2024epfv1}. Each joint query independently attends to sampled image features via learned offsets and attention weights. Outputs from different views are sequentially fused using an MLP. This design introduces the specialized component with higher development complexity~\cite{yang2024widthformer} and suboptimal hardware utilization. B, J, C, $\text{num}_\text{pts}$, and h denote batch size, number of joints, feature channels, reference points, and attention heads, respectively. Learnable layers are highlighted in \textcolor{BrickRed}{red}, with layer names matching Tab.~\ref{tab:efficiency_epfv1_compare}.}
    \label{fig:cross_att_epfv1}
\end{center}%
}]

\begin{figure}
    \centering
    \includegraphics[width=0.98\linewidth]{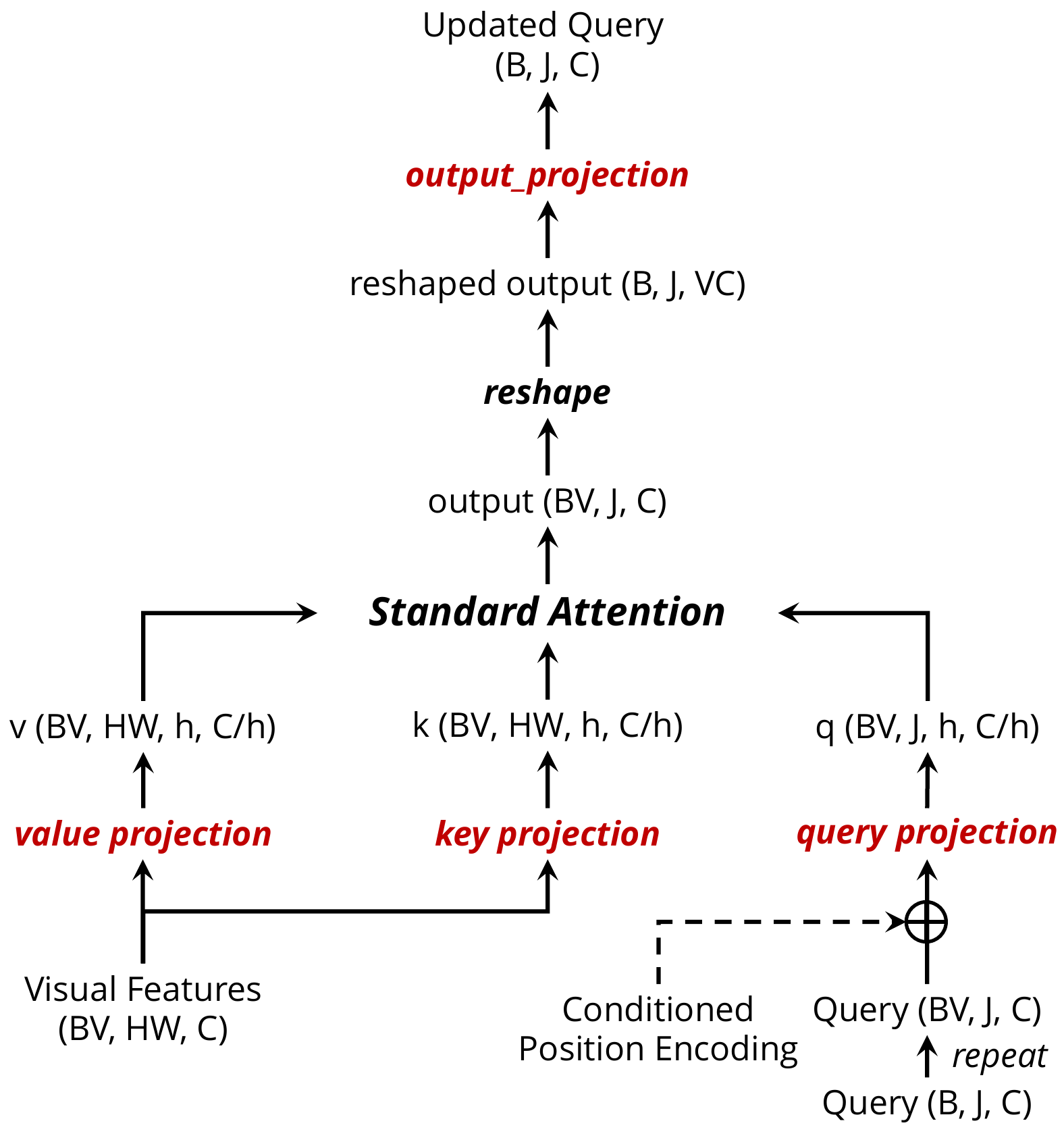}
    \captionof{figure}{Our simplified multi-view cross-attention module built on standard attention. A single holistic query attends to all view features using conditioned positional encoding in a batch manner, enabling more efficient and scalable spatial fusion. V denotes the number of views. Learnable layers are highlighted in \textcolor{BrickRed}{red}, with naming aligned to Tab.~\ref{tab:efficiency_epfv1_compare}.}
    \label{fig:cross_att_ours}
\end{figure}

\begin{table}[t!]
    \centering
    \caption{Layer-wise comparison of parameter count and FLOPs for different spatial attention modules. (1) EPFv1 uses deformable attention with 16 joint queries and reference points, introducing moderate compute and high development complexity. (2) A baseline standard attention setup with 16 queries shows 4.75$\times$ higher FLOPs due to dense spatial interactions. (3) Our final design adopts a single holistic query and standard attention, achieving the lowest computation (131K FLOPs) while maintaining similar parameter count to other variants.}
    \label{tab:efficiency_epfv1_compare}
    \scalebox{0.65}{
    \begin{tabular}{C{4cm}|C{3cm}|C{2cm}|C{2cm}}
        \toprule
        Method & Layer & \# of \textit{params} & FLOPs \\
        \midrule
        \multirow{6}{*}{\makecell[c]{(1) EPFv1 \\Deformable Attention\\w/ 16 Queries\\w/ 16 reference points}} & sampling offsets & 16512 & 32768 \\
         & attention weights  & 8256 & 16384 \\
         & value projection  & 16512 & 32768 \\
         & output projection & 16512 & 32768 \\
         & fuse mlp  & 32896 & 32768 \\
         & norm & 256 & 0 \\
        \midrule
        Total & - & 90944 & 147456 \\
        \midrule
        \multirow{5}{*}{\makecell[c]{(2) EPFv2 w/ 16 Queries\\Standard Attention}} & query projection & 16512 & 32768 \\
         & key projection  & 16512 & 32768 \\
         & value projection  & 16512 & 32768 \\
         & output projection & 32896 & 524288 \\
         & norm & 256 & 0 \\
        \midrule
        Total & - & 82688 & 622592 \\
        \midrule
        \multirow{5}{*}{\makecell[c]{(3) EPFv2 w/ 1 Query\\Standard Attention\\\textbf{(Our Final Design)}}} & query projection & 16512 & 32768 \\
         & key projection  & 16512 & 32768 \\
         & value projection  & 16512 & 32768 \\
         & output projection & 32896 & 32768 \\
         & norm & 256 & 0 \\
        \midrule
        Total & - & 82688 & 131072 \\
        \bottomrule
    \end{tabular}
    }
\end{table}

\section{Efficiency Analysis}

This section provides a detailed efficiency comparison between EPFv2 and other models like EgoPoseFormer (EPFv1)~\cite{yang2024epfv1} and EgoBody3M~\cite{zhao2024egobody3m}.

\subsection{Cross-attention in EPFv1 and EPFv2}

The key architectural difference lies in how spatial features are aggregated from multi-view images. 
EPFv1 uses deformable attention, which dynamically samples image features around projected keypoint locations using learned offsets. 
While deformable attention benefits from optimized CUDA kernels during training, achieving similar efficiency during on-device inference would incure a large amount of random memory reads~\cite{yang2024widthformer}, limiting its efficiency on edge-computing devices, \ie, the AR/VR headset in our case.
In contrast, \ourmethod employs standard cross-attention conditioned on concatenated projected 2D keypoints (Eq. 7 in the main paper), encoding multi-view information through the conditioning mechanism rather than explicit deformable sampling.
We visualize both architectures in Fig.~\ref{fig:cross_att_epfv1} and Fig.~\ref{fig:cross_att_ours}, with red labels denoting trainable components. 
Corresponding FLOPs and parameter breakdowns are summarized in Tab.~\ref{tab:efficiency_epfv1_compare}. 
Apart from its simplicity, our method achieves even smaller cost (in terms of FLOPs and parameters) while offering significantly better deployment friendliness because of its reliance on standard operations.

\subsection{The Impact of Joint Count}

An efficiency bottleneck in prior transformer-based models such as EPFv1 is the one-to-one correspondence between body keypoint and query tokens. 
Specifically, the model uses a separate query per keypoint and each query independently attends to the feature maps. 
This design introduces computational overhead that scales linearly with the number of keypoints.

In contrast, our method adopts a single holistic pose query that aggregates all necessary information. 
This design decouples the model's computational complexity from the number or type of predicted joints. 
As a result, our architecture improves inference efficiency while also enabling seamless support for different parametric body models.

Tab.~\ref{tab:efficiency_epfv1_compare} quantifies the benefits of this design. 
Compared to a 16-query baseline using standard attention, our single-query variant reduces FLOPs by over 4$\times$ (622K vs. 131K) while maintaining the same number of parameters. 
The reduced computational cost makes the model more suitable for deployment on edge devices, and the unified query design simplifies implementation and integration across different body models.

\begin{table}[t]
    \centering
    \caption{Efficiency Comparison with EgoBody3M. We compare only the pose estimation heads, assuming a shared backbone. \ourmethod significantly reduces parameter count and FLOPs, demonstrating the efficiency of its streamlined transformer design.}
    \label{tab:efficiency_egobody3m_compare}
    \scalebox{0.8}{
    \begin{tabular}{c|cc}
        \toprule
        Method & \# of \textit{params} & FLOPs \\
        \midrule
        EgoBody3M~\cite{zhao2024egobody3m} & 14.96M & 39.76G \\
        \ourmethod & 0.83M & 10.52G \\ 
        \bottomrule
    \end{tabular}}
\end{table}

\begin{table}[h]
    \centering
    \caption{
    Latency measurement of \ourmethod.
    } \vspace{-0.2cm}
    \scalebox{0.55}{
    \begin{tabular}{C{2cm}|C{0.8cm}|*{2}{C{2.2cm}}*{2}{C{2.2cm}}}
        \hline
        Backbone & FP16 & Plain PyTorch & ONNX CPU & TensorRT \textit{lvl.} 0 & TensorRT \textit{lvl.} 5  \\
        \hline
        \multirow{2}{*}{ResNet18} &  & 9.8 ms & 59.2 ms & 3.3ms & 1.3ms \\
         & \cmark & 8.3 ms & -- & 1.6ms & 1.0ms \\
        \hline
        \multirow{2}{*}{MobileNetV4S} & & 14.9 ms & 23.8 ms & 3.0ms & 1.1ms \\
         &\cmark & 13.3 ms & -- & 1.3ms & 0.8ms \\
        \hline
    \end{tabular}} 
    \vspace{-0.2cm}
    \label{tab:rebuttal:inference_speed}
\end{table}

\subsection{Efficiency Comparison with EgoBody3M}

\ourmethod and EgoBody3M~\cite{zhao2024egobody3m} are both temporal models, enabling a fair, end-to-end efficiency comparison. Here, we focus the comparison on the pose estimation head as the image encoders are subject to change.
As shown in Tab.~\ref{tab:efficiency_egobody3m_compare}, our transformer-based design significantly reduces the parameter count from 14.96M to 0.83M and FLOPs from 39.76G to 10.52G. 
This reduction in parameters and drop in FLOPs highlights the effectiveness of our proposed architecture, making it highly suitable for real-time deployment on resource-constrained devices. 

\subsection{Latency measurement}
We provide the latency measurement of \ourmethod. Specifically, we report latency for both plain PyTorch models (common in research) and ONNX models (suitable for deployment). For ONNX, CPU latency was measured using FP32 with \texttt{onnxruntime}, and GPU latency of both FP32 and FP16 were measured using TensorRT's \texttt{trtexec} with CudaGraph enabled at optimization \textit{lvl.} 0 (least optimized) and \textit{lvl.} 5 (most optimized). All experiments use 4-view 256×320 images (our baseline setting) on an Intel Xeon Platinum 8339HC CPU and NVIDIA A100 GPU.

\begin{table}[t]
    \centering
    \caption{\textbf{Comparison on Ego4View-Syn~\cite{akada2025egorear}.} We evaluate a single-frame variant of \ourmethod on the Ego4View-Syn dataset. Despite removing temporal modeling, our method achieves strong results, outperforming prior works in PA-MPJPE and remaining competitive in MPJPE.}
    \label{tab:egorear}
    \scalebox{0.8}{
    \begin{tabular}{C{3cm}|C{2cm}C{2cm}}
        \toprule
        Method & MPJPE & PA-MPJPE  \\
        \midrule
        EgoPoseFormer~\cite{yang2024epfv1} & 27.36 & 23.31 \\
        EgoRear~\cite{akada2025egorear} & \textbf{27.04} & 23.18 \\
        \ourmethod & 27.94 & \textbf{22.53} \\ 
        \bottomrule
    \end{tabular}
    }
\end{table}

\section{Additional Benchmark on Ego4View-Syn}

In the main paper, we focus our evaluation and ablation studies on the EgoBody3M~\cite{zhao2024egobody3m} dataset, since it contains long egocentric video sequences with 3D ground-truth poses. 
To further validate our architecture, we conduct an additional benchmark on the Ego4View-Syn dataset proposed in the recent SoTA method EgoRear~\cite{akada2025egorear}.

For a fair comparison with existing single-frame methods on this dataset, we adapt our model by removing the causal temporal attention module, effectively reducing \ourmethod to a single-frame baseline method. 
Following EgoRear's architecture, we use three transformer layers: one for pose proposal and two for refinement.
Since Ego4View-Syn does not provide auxiliary inputs (e.g., headset pose), we initialize the holistic query using an MLP over the image features, following EPFv1~\cite{yang2024epfv1}.

All training implementations are kept consistent with EgoRear. As shown in Tab.~\ref{tab:egorear}, our ``simplified'' single-frame version achieves competitive MPJPE and the best PA-MPJPE among all compared methods. This confirms that our architectural design offers strong performance even in the absence of temporal modeling.

\section{Additional Ablation Study}

We present two additional ablation studies to further investigate the impact of temporal context length and image backbone capacity on the performance of \ourmethod.

\begin{table}[t!]
    \caption{Impact of temporal sequence length. Using only two frames leads to significant performance degradation, while longer sequences offer stable and improved accuracy. We use 16 as the default setting for fair comparison with prior work~\cite{zhao2024egobody3m}.}
    \label{tab:seq_len}
    \centering
    \scalebox{0.85}{
    \begin{tabular}{L{2.2cm}|C{3.5cm}}
        \toprule
        Length & Overall MPJPE \\
        \midrule
        2 & 4.30 \\
        16 \textbf{(default)} & \textbf{4.17} \\
        \bottomrule
    \end{tabular}
    }
    \vspace{-5pt}
\end{table}

\subsection{Sequence Length}

We vary the causal temporal attention window length to assess the contribution of temporal modeling.
As shown in Tab.~\ref{tab:seq_len}, using only two frames leads to a substantial performance drop, indicating that temporal context is crucial for accurate 3D pose estimation. 
For longer sequences, performance remains stable. 
We adopt a length of 16 frames as the default setting, following EgoBody3M~\cite{zhao2024egobody3m}, to ensure both fair comparison and efficient training.

\begin{table}[t!]
    \centering
    \caption{Comparison across different image backbones. Larger backbones, such as DINOv3-B, significantly improve pose estimation accuracy, demonstrating the benefit of strong visual features in egocentric settings.}
    \label{tab:variant_backbone}
    \scalebox{0.85}{
    \begin{tabular}{L{3cm}|C{2.5cm}|C{2.5cm}}
        \toprule
        Variants & \# of \textit{params.} & Overall MPJPE \\
        \midrule
        ResNet-18~\cite{he2016resnet} & 12.5M &  4.17 \\
        ResNet-50~\cite{he2016resnet} & 43.3M &  4.03 \\
        DINOv3-ViT-B~\cite{simeoni2025dinov3} & 119.2M & \textbf{3.96}  \\
        \bottomrule
    \end{tabular}
    }
\end{table}

\subsection{Image Encoders}

We also evaluate the effect of different image backbone architectures, ranging from lightweight (ResNet-18~\cite{he2016resnet}) to large-scale vision transformers (DINOv3-B~\cite{simeoni2025dinov3}). Results in Tab.~\ref{tab:variant_backbone} show a clear trend: stronger backbones consistently yield better 3D pose estimation, with DINOv3-B achieving the best performance, highlighting the benefit of high-quality features for challenging egocentric scenarios.

\end{document}